# CRATOS: Cognition of Reliable Algorithm for Time-series Optimal Solution

Ziling Wu
School of Electronics and Information Technology,Sun Yat-sen University
Xiaoguwei Island, Panyu District
Guangzhou, 510006, P. R. China
wuzling@mail2.sysu.edu.cn

Ping Liu
Huawei Technologies Co. Bantian, Longgang District
Shenzhen 518129, P. R. China
liuping29@huawei.com

Zheng Hu
Huawei Technologies Co. Bantian, Longgang District
Shenzhen 518129, P. R. China
hu.zheng@huawei.com

Bocheng Li
School of Electronics and Information Technology,Sun Yat-sen University
Xiaoguwei Island, Panyu District
Guangzhou, 510006, P. R. China
libch8@mail2.sysu.edu.cn

Jun Wang
School of Microelectronics Science and Technology,Sun Yat-sen University
Tangjiawan village,Xiangzhou district
Zhuhai, 519082, P. R. China
wangj387@mail.sysu.edu.cn

## ABSTRACT
Anomaly detection of time series plays an important role in reliability systems engineering. However, in practical application, there is no precisely defined boundary between normal and anomalous behaviors in different application scenarios. Therefore, different anomaly detection algorithms and processes ought to be adopted for time series in different situation. Although such strategy improve the accuracy of anomaly detection, it takes a lot of time for practitioners to configure various algorithms to millions of series, which greatly increases the development and maintenance cost of anomaly detection processes. In this paper, we propose CRATOS which is a self-adapt algorithms that extract features from time series, and then cluster series with similar features into one group. For each group we utilize evolutionary algorithm to search the best anomaly detection methods and processes. Our methods can significantly reduce the cost of development and maintenance of anomaly detection. According to experiments, our clustering methods achieves the state-of-art results. The accuracy of the anomaly detection algorithms in this paper is 85.1%.

## 1. INTRODUCTION
Outlier detection has become a field of interest for many researchers and practitioners and is now one of the main tasks of time series data mining. Although various anomaly detection algorithms have been investigated, there is no universal algorithm to deal with all the anomaly detection tasks in time series. In most cases, appropriate anomaly detection algorithms are manually configured to a KPI (Key Performance Indicator), such as CPU utilization or queries per second that arrive as time series, based on the nature of it, which is helpful to supervise whether a system works normally. Especially, the inappropriate adaptation which frequently happens will lead to serious problems such as false positive, false negative and untimely alarm. Moreover, time series in cloud computing data centers is considerably huge. Therefore, it is pretty tedious for engineers to manually adapt the detection process for different time series.[16].

In order to solve the problem mentioned above, Li et al.[16] proposed a KPIs clustering method in which the complicated time series is gathered into one cluster according to similarities, and then uniform anomaly detection algorithm is adapted to these clustered time series. However, most existing clustering methods based on the waveform similarity hardly take use of this idea because such clustering methods are not designed for anomaly detection on purpose. The clustering results gotten from those methods are sometimes not suitable for any anomaly detector.

In this paper, we propose CRATOS which can self-adapt algorithms for anomaly detection, as shown in Figure 1. Firstly, the features of a large number of input time series are extracted, and then targeted hierarchical clustering for the extracted features are made. Then evolutionary algorithm (EA) is used to find the most appropriate anomaly detection algorithm and corresponding parameters for each cluster. After the off-line training process above, a trained anomaly detection mode can be obtained as: a) determine the cluster of the input time series; b) use the previously trained anomaly detection process which is suitable for the cluster that the input time series belongs to to detect outliers in the input time series. The above processes can significantly improve the efficiency of algorithm adaptation and maintenance for large-scale time series anomaly detection.

In summary, the contribution of the paper consists of:

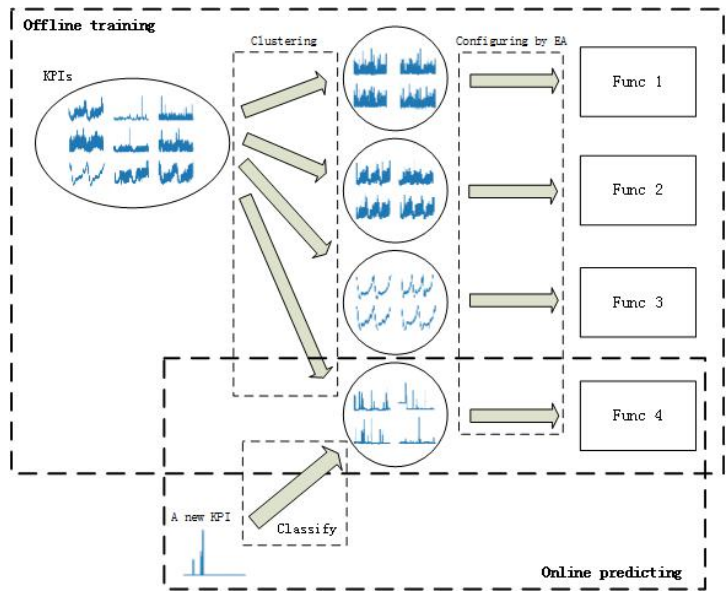

**Figure 1.** The overview of our frame work. In the section of offline training, we first cluster KPIs in to a number of clusters, then we configure parameters and algorithms for each cluster with evolution algorithm. In the section of online predicting, we determine which cluster a new KPI belongs to and then we use the anomaly detection processes for the cluster to detect outliers in the new KPI.

1) three features is proposed for clustering and suggest to cluster time series with these features.

2) evolutionary algorithms is utilized to select appropriate anomaly detection algorithms and parameters for different KPIs' clusters.

3) We suggest a self-adapt algorithm processes for anomaly detection, which we name CRATOS.

## 2. RELATED WORKS

There are various anomaly detector for different kinds of detection problems. As for points outliers in univariate time series, one strategy is to predict the expected value of a point and then compare the difference between the ground truth value of that point and the expected value. That strategy is widely used in [6, 7, 22, 27, 4, 10, 2]. In terms of multivariate time series, the anomaly detection methods of single KPI was applied to multivariate time series after eliminating the correlation between different KPIs, such as [21, 9, 3, 18, 26, 23, 15, 28, 29].

When it comes to subsequence outliers of univariate time series, Keogh et al. [14] and Lin et al. [17] compare the difference between a subsequence and the other subsequences in the time series. However, such strategy requires an artificially predetermined length of the subsequence. To address this problem, Senin et al. [25] utilize Piecewise Aggregate Approximation (PAA) to calculate the length of subsequences automatically.

For subsequence outliers in multivariate time series, compared with univariate time series, it is often necessary to consider the correlation between KPIs. Jones et al. [12, 13] and Wang et al. [30] directly apply univariate techniques to each time-dependent variable in multivariate time series. Different from that, Munir et al. [19] use Convolutional Neural Network to predict the expected value of the upcoming subsequence and compare it with the ground truth value to detect outlier subsequence.

In recent years, deep learning methods including [11, 24, 31] has been widely used in the field of anomaly detection. However, The boundary between abnormal and normal behaviors is not clear defined in different data domains and there is no enough labeled datasets for supervised learning. Although it has better performance than traditional methods, deep learning cannot replace the traditional methods in anomaly detection.

Since boundaries between normal and abnormal behaviors are not consistent in different application scenarios, even though many state-of-art anomaly detection algorithms have been proposed, we still need to configure different anomaly detection techniques to different application scenarios. One of the solutions to this problem is to divide a large number of KPIs into different clusters, and the KPIs in a single cluster have much in common. Then, a unified anomaly detection algorithm can be adapted to a single cluster, so as to avoid configuring anomaly detection algorithms to a large number of KPIs one by one. This idea requires us to make a classifier of massive KPIs and a matching strategy of algorithms. As for the KPIs classifier, the application of supervised learning method is not effective since there is no clear classification of KPIs categories in the industry and there is also a lack of labeled data sets for KPIs classification. Therefore, many scholars have focused on clustering, which is an unsupervised learning strategy. Ding et al. [8] proposed YADING which can cluster large scale time series. By randomly extracting time series and using PAA to compress time series, it can reduce the computation quantities of large scale KPIs clustering. By taking L1-norm between compressed time series as the distance of time series, it cluster compressed time series with multi-DBSCAN. Li et al. [16] proposed ROCKA which can also cluster large scale time series. Similar to YADING, ROCKA also uses DBSCAN for clustering. The difference is that ROCKA uses shape-based distance (SBD) distance to measure the distance between time series. However, both ROCKA and YADING have drawbacks. The PAA used by YADING is essentially mean filtering with strides bigger than 1, and the series that are finally clustered are the smoothed series.

ROCKA removes impulses and extracts baselines through mean smoothing before clustering. Therefore, both YADING and ROCKA cannot distinguish the amplitude and impulses density of KPIs. In addition, it is inevitable to identify some KPIs as noise with DBSCAN, to be more specific, these KPIs do not belong to any cluster, which leads to insufficient use of data.

In terms of configuring strategies of algorithms, Bergstra et al. [5] suggest that the traditional search strategy (grid search) is often less effective than the random search because there are many dimensions in the search space that have little impact on the results, and it is difficult to determine which of these dimensions have little impact on a specific problem but grid search wastes a lot of time searching on these useless dimensions. In [1], genetic programming is utilized to search for the most suitable algorithm for different specific problems.

In summary, anomaly detection algorithms based on data analysis has been developed and applied in industry for a long time. However, there are still some problems in the self-adapt algorithm of millions of time series data in the cloud scenario even though the industry has some good practices such as YADING and ROCKA. Automatic configuring algorithms is also a problem to be solved, the industry is just starting on this issue.

## 3. PROPOSALS

Similar interval tendency, amplitude, impulses density are the main factors that we consider most when we configure algorithms for a KPI. For example, dynamic threshold method makes sense when dealing with time series with periodicity or similar interval tendency as shown in Figure 2(a)(b). On the other hand, it is not necessary to consider historical trends when dealing with nonperiodic signals as shown in Figure 2(c)(d). Besides, as shown in Figure 2(a)(b), since the amplitude of the two curves is different due to the noise, different tolerances are needed to detect steep drops. In addition, there are always sparse impulses in time series, as shown in Figure 2(c), such waveforms are suitable for median smoothing in the pre-processing step because these sparse impulses are not abnormal and are caused by many transient behaviors in practice, such as system jamming and JVM garbage collection. On the contrary, for the case shown in Figure 2(d), these dense impulses are the normal state of the system instead of anomalies, so it is necessary to use mean smoothing in the pre-processing step. Therefore, different anomaly detection algorithms and detection processes should be adapted to time series with different properties.

In this section, we propose a framework from clustering to matching appropriate detection algorithms for different clusters of time series, as shown in Figure 1.

### 3.1 Clustering

In order to ensure that the clustering result can be a perfect match corresponding anomaly detection algorithm, we hope that the clustering results can distinguish time series significantly in the three properties of periodicity, amplitude and impulse density.

Therefore, it is an effective way to extract different features for these three properties and conduct targeted hierarchical clustering. In this section, we will expound the features required for hierarchical clustering.

#### 3.1.1 Clustering methodology
Both ROCKA and YADING choose DBSCAN as the clustering method. The reason is that the number of clusters is uncertain.

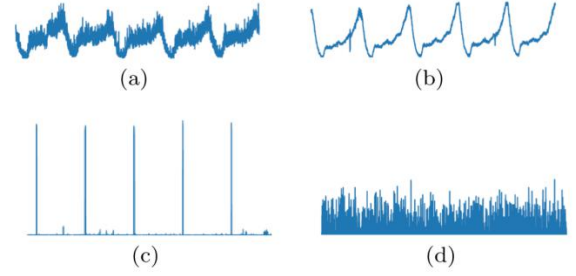

Figure 2: KPIs with different properties. KPIs with interval tendency and large or small amplitude (a) (b). KPIs without interval tendency and with sparse or dense impulses (c) (d).

However, hierarchical clustering is utilized to classify the categories according to the periodicity, amplitude and impulse density, so the number of clusters are very clear. Therefore, k-means is suitable for clustering. And we set k to 2 for each layer of clustering, where k is the number of clusters.

#### 3.1.2 Section-sign feature
Inspired by local binary patterns (LBP) [20] which is widely used in digital image processing and computer vision, a new feature named Section Sign is proposed. Compared with applying LBP directly to time series, our Section-sign feature is faster to calculate and can effectively distinguish the time series with or without periodicity.

A. Pre-processing

Before extracting Section-sign features, we should remove impulses from the data. We recommend taking 1.01 times the 99 percentile in a single time series as the upper bound, and 0.99 times the one percentile as the lower bound, replacing all values greater than the upper bound with the upper bound, and all values less than the lower bound with the lower bound.

B. Extracting Section-sign features of KPIs

The set composed of multiple time series after pre-processing is defined as $T = \{T_1, T_2, T_3, ..., T_n\}$. A time series of length $l$ is defined as $T_i = [T_{i1}, T_{i2}, T_{i3}, ..., T_{il}]$, $T_i \in T$, $i \in [1, n]$. In order to extract the Section-sign feature of $T_i$, we use a slide-window with length $m = 90$ to slide from the beginning of $T_i$ to the end of $T_i$ with the stride $s = 30$. A time series segment covered by the slide-window is defined as:

$t_{ij} = [T_{i[(j-1)s+1]}, T_{i[(j-1)s+2]}, T_{i[(j-1)s+3]}, ..., T_{i[(j-1)s+m]}]$, where

$j \in [1, \lfloor \frac{l-m}{s} + 1 \rfloor]$. With each sliding step, the Section-sign features of $t_{ij}$ is extracted. The method is as follows:

1) define *med* as the value of the center point of $t_{ij}$. *med* is calculated by equation 1:

$$med = \begin{cases} t_{ij}[\lceil \frac{m}{2} \rceil], & m > 1 \text{ and } m \text{ is odd} \\ \frac{t_{ij}[\frac{m}{2}] + t_{ij}[\frac{m}{2}+1]}{2}, & m > 1 \text{ and } m \text{ is even} \end{cases} \qquad (1)$$

2) calculate *diff* which is defined as: $diff = [t_{ij}[1] - med, t_{ij}[2] - med, ..., t_{ij}[m] - med]$, then *diff* is divided into two

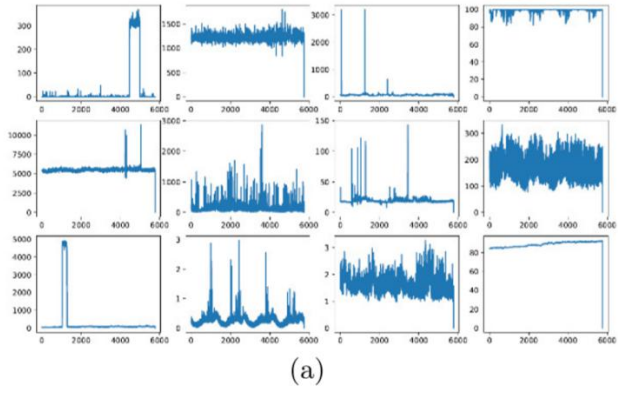
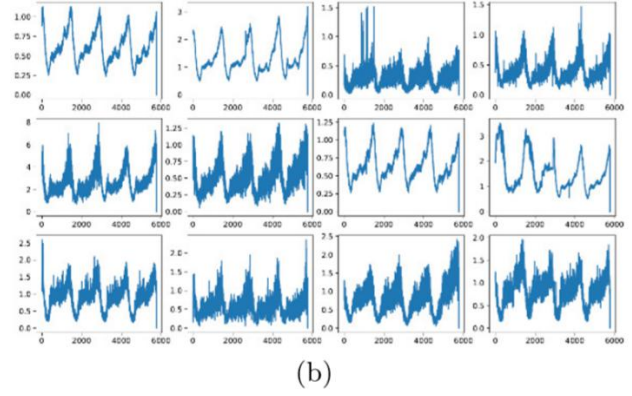

Figure 3: The results of clustering with Section-sign feature. KPIs without periodicity or similar interval tendency (a). KPIs with periodicity or similar interval tendency (b).

parts of the same length. The left half is defined as *left*, and the right half is *right*.

3) The Section-sign feature of $t_{ij}$ is defined as: $S_{ij},S_{ij+1} = mean(sign(left)), mean(sign(right))$. The Section-sign feature of $T_i$ is defined as: $S_i = [S_{i1},S_{i2},S_{i3},S_{i4},...,S_{i(2h)}]$, where $h = \lfloor\frac{l-m}{s} + 1\rfloor$. *mean* is to calculate mean value, *sign* is defined in equation 2:

$$sign(x) = \begin{cases} 1, & x > 0 \\ 0, & x = 0 \\ -1, & x < 0 \end{cases} \quad (2)$$

Intuitively, both Section-sign feature and LBP are looking for trends within a segment. Therefore, the Section-sign features can effectively extract the difference between cyclical trends and aperiodic trends to classify time series. After extracting the feature sequence $S_i$ of all $T_i$ in $T$, we utilize $k-means$ to cluster $T$ and obtain the periodic sequence set $T\_1$, the non-periodic sequence set $T\_0$. The results of clustering with Section-sign feature is shown in Figure 3.

### 3.1.3 Swing feature

To distinguish the amplitude of time series, Swing feature is proposed. Time series set T has already been divided into two parts $T\_1$ and $T\_0$ before extracting Swing feature. Swing feature can distinguish the amplitude of $T\_1$ and $T\_0$, respectively. It is worth noticing that when we extract features, the data pre-processing processes of different features are different and independent of each other because the information that messes the extraction result of a feature may be the key information desired by other features. For instance, the impulses and Gaussian noises which affect the result of Section-sign feature extraction and need to be filtered out are important factors deciding the amplitude of time series and should be retained. The result of distinguishing the amplitude of $T\_1$ from $T\_0$ is shown in Figure 4.

A. Pre-processing

The pre-processing of Swing features also requires the removal of up and down impulses from the data, in a manner consistent with Section-sign features, which is followed by maximum-minimum normalization.

B. Extracting Swing features of KPIs

Similar to Section-sign feature, we define the set composed of multiple time series after pre-processing as $T = \{T_1,T_2,T_3,...,T_n\}$. A time series of length $l$ is defined as $T_i = [T_{i1},T_{i2},T_{i3},...,T_{il}]$, $T_i \in T$, $i \in [1,n]$. Then we calculate the first-order difference of $T_i$ as $D_i = [D_{i1},D_{i2},D_{i3},...,D_{il-1}]$. We use a slide-window with length $m = 90$ to slide from the beginning of $D_i$ to the end of $D_i$ with the stride $s = 30$. A time series segment covered by the slide-window is defined as:

$d_{ij} = [D_{i[(j-1)s+1]},D_{i[(j-1)s+2]},D_{i[(j-1)s+3]},...,D_{i[(j-1)s+m]}]$, where

$j \in [1,\lfloor\frac{l-1-m}{s}+1\rfloor]$. With each sliding step, Swing features of $d_{ij}$ is extracted, The method is as follows:

1) calculate the deference between the $80^{th}$ and $20^{th}$ percentiles of $d_{ij}$ as $w$.

2) define the Swing feature of $d_{ij}$ as $w_{ij} = w$, and the Swing feature of $T_i$ is: $W_i = [w_{i1},w_{i2},w_{i3},w_{i4},...,w_{i(h)}]$, $h = \lfloor\frac{l-1-m}{s}+1\rfloor$.

It's easy to understand that Swing feature is similar to variance, essentially measuring the volatility of a time series to determine the amplitude.

### 3.1.4 Diff-Thres feature

The Diff-Thres feature is proposed to classify the density of impulses in time series. It is located in the last step of hierarchical clustering. The sequence sets after Swing feature classification are respectively clustered. The clustering results is shown in Figure 5.

It's not necessary to pre-process the data before extracting Diff-Thres feature because impulses whose density is what we care about shouldn't be filtered. We define the set composed of raw multiple time series as $T = \{T_1,T_2,T_3,...,T_n\}$. A time series of length $l$ is defined as $T_i = [T_{i1},T_{i2},T_{i3},...,T_{il}]$, $T_i \in T$, $i \in [1,n]$. Then we calculate the first-order difference of $T_i$ and take the absolute value as $D_i = [D_{i1},D_{i2},D_{i3},...,D_{il-1}]$. We use a slide-window with length $m = 180$ to slide from the beginning of $D_i$ to the end of $D_i$ with the stride $s = 30$. A time series segment covered by the slide-window is defined as:

$d_{ij} = [D_{i[(j-1)s+1]}, D_{i[(j-1)s+2]}, D_{i[(j-1)s+3]}, ..., D_{i[(j-1)s+m]}]$, where $j \in [1, \lfloor \frac{l-1-m}{s} + 1 \rfloor]$. With each sliding step, The Diff-Thres features of $d_{ij}$ is extracted, The method is as follows:

1) define $max$ as the maximum value of $d_{ij}$.

2) set three attenuation coefficients $div = [2,3,4]$ and calculate $threshold = [\frac{max}{2}, \frac{max}{3}, \frac{max}{4}]$.

3) for $d_{ijk} \in d_{ij}$, where $k \in [2, m-2]$, when $d_{ijk-1} < threshold < d_{ijk}$ or $d_{ijk} < threshold < d_{ijk-1}$ happens, it is considered that a impulse has occurred.

The specific process is shown in Table 1.

## 3.2 Choosing anomaly detection algorithms and parameters for each cluster

After obtaining the results of the hierarchical clustering, we need to configure the anomaly detection process for each cluster, including pre-processing for time series such as normalization, smooth method, smooth window size, anomaly detector, selections of anomaly detectors and other hyper-parameters required by the anomaly detector. So it is tedious to manually configure the appropriate anomaly detection process for each cluster. Traversal search method can also lead to the problem of combination explosion. Therefore we suggest using evolutionary algorithms to select appropriate anomaly detection algorithms and parameters for different KPIs' clusters. Compared with manually configuring and traversal search, evolutionary algorithms can significantly reduce the effort spent on configuring algorithm. In this section we will discuss how to apply evolutionary algorithms to the configuration of anomaly detection processes.

### 3.2.1 Configuring the process of anomaly detection

There are many factors to consider in the design of the algorithm detection process. Sometimes it is necessary to select one of two or more functions as one step in the entire anomaly detection process. For example, mean smoothing or median smoothing is needed in data pre-processing. Sometimes we need to design the execution sequence of various methods, such as whether the detector should detect the dynamic threshold first or the steep rise and fall first. Sometimes it is necessary to determine whether a step should be carried out, for example, whether data should be normalized before testing. In a word, the design of the detection process mainly faces three problems: a) the choice of function b) the setting of the execution order of multiple functions c)the setting of whether to execute a certain step. For evolutionary algorithms, how to solve these three problems, how to initialize the values of genes, and how to design methods for variation, are the crucial problems. Other aspects of evolutionary algorithms, such as reproduction and natural selection, are not covered in this paper. Next we will discuss the above issues separately.

A. Initialization

1) Initialize functions

We can put all the functions to be selected into a set and randomly select one or more functions as the selected functions with the same probability when initialize functions.

2) Initialize sequence of multiple functions

We can put all the methods in one set and shuffle them.

3) Initialize whether a method is performed

We can randomly generate a Boolean value that is initialized to execute the method when it is True and not executed when it is False.

**Table 1. The process of extracting Diff-Thres feature of a KPI**

| | |
|---|---|
| Func 1 | Input:<br>   $T_i$: Input one single time series<br>Output:<br>   features: Diff-Thres feature of the time series<br>def get_features_for_sparse($T_i$):<br>   # calculate the first-order difference of $T_i$<br>   # and take the absolute value<br>   $D_i$=abs(diff($T_i$))<br>   cross2=get_cross_feature($D_i$,div=2)<br>   cross3=get_cross_feature($D_i$,div=3)<br>   cross4=get_cross_feature($D_i$,div=4)<br>   cross = concat([cross2,cross3,cross4], axis=1)<br>   # maximum-minimum normalization<br>   features = MinMaxScaler(cross)<br>   return features |
| Func 2 | def get_cross_feature($D_i$,div):<br>   m=180<br>   s=30<br>   cross_thres=[ ]<br>   for $d_{ij}$ in $D_i$:<br>      threshold=max($d_{ij}$)/div<br>      num=0<br>      for k in range(1,m,stride=2):<br>         if $d_{ijk-1} < threshold < d_{ijk}$ or \\<br>         $d_{ijk} < threshold < d_{ijk-1}$:<br>         num+=1<br>      cross_thres.append(num)<br>   return cross_thres |

4) Initialize a specific parameter value

For a specific value, we set its value range and data type (integer, float), and then generate it randomly. It is important that the above initialization steps can be nested within each other, for example, when multiple functions are selected from a number of alternatives, it is possible to consider the order in which the selected functions are executed. In addition, in order to prevent the parameters obtained from the evolutionary algorithms do not tally with the practical experience, even if the objective function were significantly improved, we should also manually limit the range of parameters according to practical experience before the parameters are initialized, so as to prevent the evolutionary algorithm from learning unreasonable parameter combinations and accelerate the convergence rate of the objective function during training.

5) Initialize the mutation rate

We define a real number rate as mutation rate. The value range of *rate* is related to the value of specific need of variation. When mutating the parameters that choose one or more functions from a set of functions, adjusting the execution order of multiple functions and whether a method is executed or not, $rate \in (0,1)$. As for a specific parameter value is mutated, the value range of *rate* is related to the order of magnitude of the parameter. *rate* is generated randomly during initialization. In addition, *rate* is also a mutable parameter, and its initialization strategy is consistent with the initialization strategy for a specific parameter value. It is worth noticing that the *rate* of *rate* cannot be mutated and needs to be set artificially.

B. Mutation

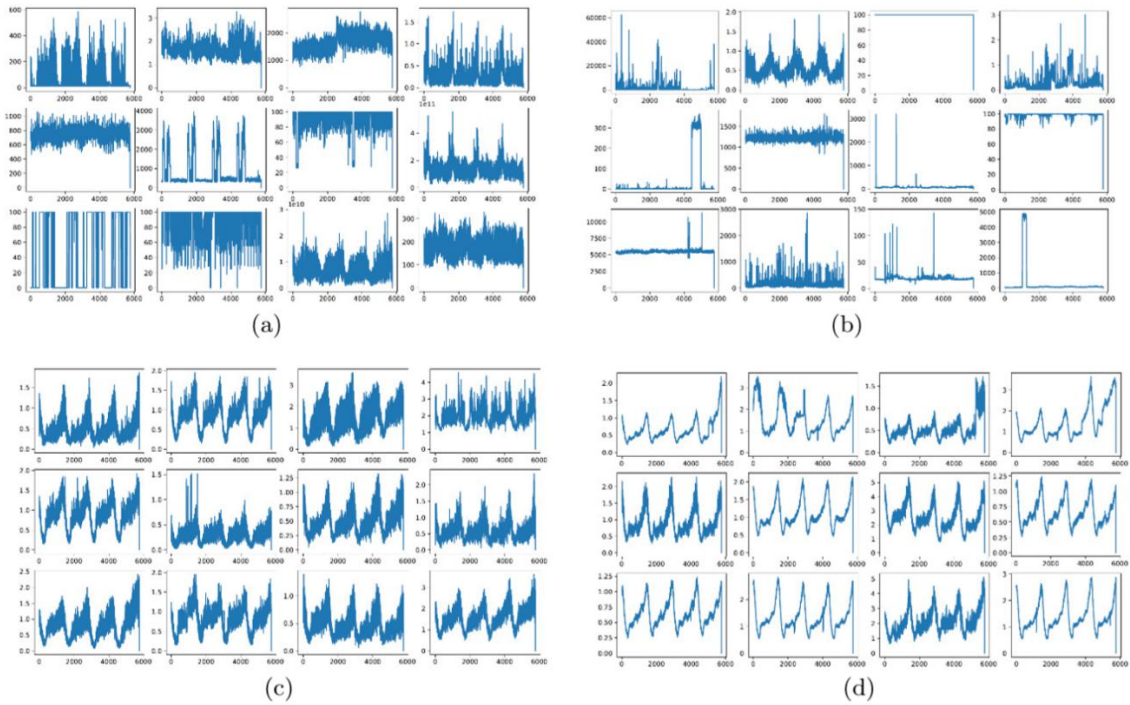

**Figure 4: The results of clustering with Swing feature. KPIs with large amplitudes and without similar interval tendency (a). KPIs with small amplitudes and without similar interval tendency (b). KPIs with large amplitudes and similar interval tendency (c). KPIs with small amplitudes and similar interval tendency (d).**

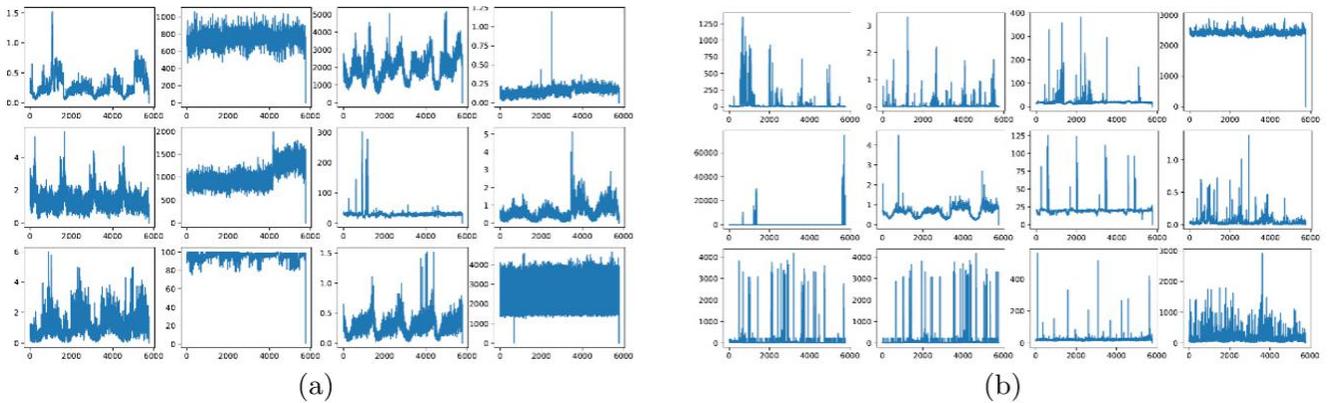

**Figure 5: The results of clustering with DiffThres feature. KPIs with dense impulses (a). KPIs with sparse impulses (b).**

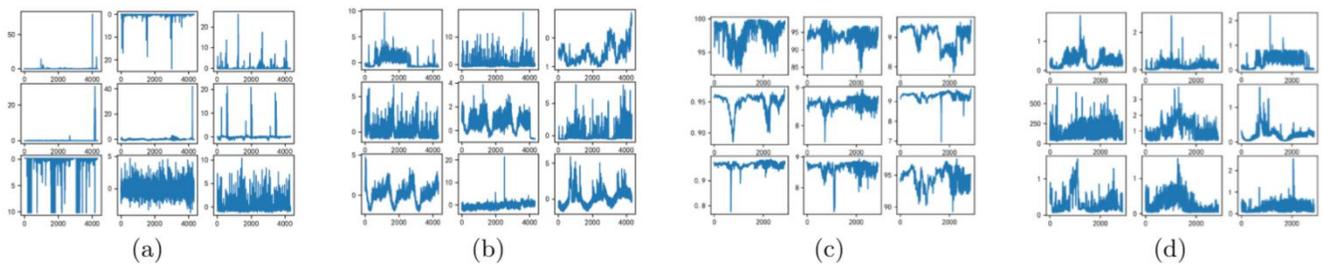

**Figure 6: Clusters by YADING and ROCKA. One of all clusters generated by YADING (a). KPIs which is classified as noises by YADING (b). One of all clusters generated by ROCKA (c). KPIs which is classified as noises by ROCKA (d).**

For the sake of discussion, method selection, execution order, and whether to execute a function are collectively referred to as function selection.

1) The process of mutation in functions selection

Mutation in functions selection is essentially a matter of deciding whether or not to mutate, in which case a random floating-point number r is generated, $r \in (0,1)$, if $r > rate$ then mutate the functions selection. Therefore, for this kind of problems, the value range of rate should be within (0,1). When mutating, we simply re-execute the initialization process for functions selection.

2) Mutation for the value of a specific parameter

The mutation for a specific parameter value is essentially a process of generating a new random number with a probability density function, which is defined as a normal distribution model $X \sim N(\mu, \sigma^2)$, where $\mu$ is the current value of the parameter value to be mutated, and $\sigma$ is the mutation *rate* of the parameter. The normal distribution model is used as a probability density function to generate new random numbers, which can be used as the new values of the mutated parameters. The mutation strategy of *rate* should also be implemented according to this strategy.

### 3.2.2 Objective function

For different business scenarios, the objective function is different, take HUAWEI Cloud BU for instance: we obtained 206 time series from HUAWEI Cloud BU, containing different cases. Some cases have anomaly which are hardly detected in traditional method, some normal KPIs are easy to false positives. Therefore, we have developed its own passing criteria for each time series, that is, no anomaly is missed, no delay in reporting and no false alarm. For a combination of parameters, we count the number of KPIs that meet this criteria during anomaly detection process as the objective function *pass_num*. We want to maximize this objective function as the evolutionary algorithm iterates.

## 3.3 Anomaly detection

In the previous process, eight clusters is obtained through the hierarchical clustering, and then evolutionary algorithm is used to determine the anomaly detection process and some key parameters for each cluster. Finally we will get a complete end-to-end anomaly detection model. It no longer requires artificial selection of matching algorithms and parameters for a certain time series. And practitioners only need to input the new time series into the model, and obtain selected suitable anomaly detection algorithms and parameters based on the results of offline trained model. The specific process is as follows:

1) Use the trained k-means model to predict which cluster the new time series belongs to.

2) After judging the cluster of the time series, the anomaly detection algorithm and the optimal parameters of the cluster are applied to the new time series.

## 4. EXPERIMENT

## 4.1 Dataset

### 4.1.1 Simulated data

In order to simulate the business data as much as possible, we adopted the method of extracting the baseline of the business data and adding Gaussian noise and pepper and salt noise to generate the simulated data or pseudo-data. The pseudo-data was only marked for the cluster of the time series, and no artificial anomaly was added. The specific process is as follows:

1) Select several representative business data.

2) Get baseline of these chosen business data by mean smoothing and median smoothing.

3) Add different levels of noise to the baseline to generate pseudo-data that can simulate business data, and label it based on the generation process.

The situation of the dataset is described in Table 2.

### 4.1.2 Data from business scenario

We obtained 206 KPIs with the length of 5760 from HUAWEI Cloud BU and marked their clusters and outliers. The situation of the dataset is described in Table 2. It should be emphasized that there are few businesses whose KPIs belong to the three clusters of TTF, TFF and FTF in business of HUAWEI Cloud BU, so there is a certain imbalance in the samples of these three clusters.

## 4.2 Evaluation standard

In order to measure the accuracy of clustering results, three indicators is utilized: precision, recall ratio and F1-score. When it comes to evaluation standard for anomaly detection, as mentioned earlier, 206 time series containing different cases was obtained from HUAWEI Cloud BU. Some cases have anomaly which are hardly detected in traditional methods, some normal KPIs are easy to false positives. Therefore, we have developed its own passing criteria for each time series instead of the precision and recall rate of anomaly detection. To be more specific, we measure the anomaly detection accuracy by the pass rate,which is defined as follows:

$$accuracy = \frac{PassNumber}{TotalNumber} \qquad (3)$$

where PassNumber is the number of series which pass the anomaly detection test without errors, TotalNumber is the total number of time series, and accuracy is the pass rate.

## 4.3 Results

### 4.3.1 Results on a single feature

First, we clustered time series with their Section-sign feature, Swing feature and Diff-Thres feature for similar interval trend, amplitude and impulses density and measured their performance respectively, as shown in Table 3.

### 4.3.2 Comparison among different clustering methods

ROCKA and YADING were chosen as the control group. Since the samples of our business data are not balanced, our method and the performance of these two methods was tested on pseudo-data to illustrate the effectiveness of our algorithm. It is worth noticing that ROCKA and YADING are not clustering algorithms specifically for the similar interval trend, amplitude and impulses density. Therefore, for the convenience of comparison, the clustering clusters generated by these two methods were combined. The merging strategy is as follows:

Table 2. The overview of our datasets

| Similar interval trend (T is having similar interval trend) | Amplitude (T means having large amplitude) | Impulses density (T means having dense impulses) | The number of simulated data | The number of data from business scenario |
|---|---|---|---|---|
| F | F | F | 500 | 50 |
| F | F | T | 500 | 32 |
| F | T | F | 500 | 5 |
| F | T | T | 500 | 56 |
| T | F | F | 500 | 1 |
| T | F | T | 500 | 30 |
| T | T | F | 500 | 0 |
| T | T | T | 500 | 32 |

Table 3. The clustering results of k-means on a single feature

| | state | precision | recall | F1 score |
|---|---|---|---|---|
| Section-Sign | T | 0.86 | 0.77 | 0.81 |
| | F | 0.79 | 0.87 | 0.83 |
| Swing | T | 0.95 | 0.84 | 0.89 |
| | F | 0.86 | 0.96 | 0.9 |
| Diff-Thres | T | 0.87 | 0.72 | 0.79 |
| | F | 0.76 | 0.9 | 0.83 |

1) count the number of KPIs of different categories in each cluster according to the labels.

2) for a cluster, take the category with the largest number in the cluster as the category of the cluster.

3) merge clusters of the same category into the final clustering results.

The final comparison results are shown in Table 4. Besides, as shown in Figure 6(a)(c), we obtained one of all clusters generated by YADING and ROCKA. It is obvious that ROCKA is able to cluster KPIS with similar waveform. However, YADING and ROCKA inevitably identify some time series as noises due to the principle of DBSCAN, which leads to underutilization of the dataset. Figure 6(b)(d) show noises generated by YADING and ROCKA. Compared them with the clustering results of CRATOS, as shown in Figure 3, 4 and 5, we can see that CRATOS not only distinguishes the time series of different features well, but also do not identify some KPIs as noises during the clustering process.

*4.3.3 Anomaly detection results*
In this part, the data obtained from HUAWEI Cloud BU was clustered, and the evolutionary algorithm was used to configure algorithms and parameters for each cluster. The use of business data can better reflect that our self-adapt anomaly detection algorithm meets the requirements of actual business scenarios. When setting up the evolutionary algorithm, we try to let the evolutionary algorithm configure the detection process for different clusters, including:

1) detector: dynamic threshold detector, global threshold detector, local steep drop detector, global steep drop detector and other detectors

2) smoothing: mean smoothing and median smoothing

3) some settings about parameter: smooth window size, sensitivity, etc

Table 4. Performances of YADING, ROCKA and our method

| | | YADING | ROCKA | CRATOS |
|---|---|---|---|---|
| FFF | precision | 0.413 | 0.374 | **0.649** |
| | recall | 0.422 | 0.432 | **0.742** |
| | F1_score | 0.417 | 0.401 | **0.692** |
| FFT | precision | 0.381 | 0.381 | **0.637** |
| | recall | 0.278 | 0.312 | **0.782** |
| | F1_score | 0.321 | 0.343 | **0.702** |
| FTF | precision | 0.366 | 0.351 | **0.739** |
| | recall | 0.366 | 0.368 | **0.748** |
| | F1_score | 0.366 | 0.359 | **0.743** |
| FTT | precision | 0.377 | 0.377 | **0.733** |
| | recall | 0.326 | 0.356 | **0.742** |
| | F1_score | 0.349 | 0.366 | **0.737** |
| TFF | precision | 0.316 | 0.274 | **0.751** |
| | recall | 0.238 | 0.78 | **0.782** |
| | F1_score | 0.271 | 0.406 | **0.766** |
| TFT | precision | 0.32 | 0.317 | **0.717** |
| | recall | 0.114 | 0.154 | **0.76** |
| | F1_score | 0.168 | 0.207 | **0.738** |
| TTF | precision | 0.343 | 0.338 | **0.863** |
| | recall | 0.114 | 0.098 | **0.694** |
| | F1_score | 0.171 | 0.152 | **0.769** |
| TTT | precision | 0.256 | 0.369 | **0.886** |
| | recall | 0.716 | 0.138 | **0.622** |
| | F1_score | 0.377 | 0.201 | **0.731** |

In addition, we initialize a population composed of different gene combinations with the scale of 200. After each iteration, only 40 gene combinations with the best performance were retained, and 160 offspring were propagated in the next iteration to control the population to remain at 200 with the purpose of controlling the computational complexity of the evolutionary algorithm at $O(n)$, which means the population size is positively correlated with the computational resources required by the evolutionary algorithm.

A server is used to train evolutionary algorithms and accelerated the training process in a multi-processes manner. In the experiment, each subprocess tested the pass rate of a gene combination for the entire time series. We made 100 sub-processes work at the same time. It took two hours for each iteration to compute. Generally, the solution of the optimal pass rate can be obtained after about 9 iterations. To ensure convergence, we iterated 40 times in total. The final pass rate of the detection process configured by the evolutionary algorithm for each cluster is 0.851.

## 5. CONCLUSION

In this paper, an KPIs clustering algorithm is proposed to conduct targeted hierarchical clustering for the features, including similiar interval tendency, amplitude, and impulses density, required by the anomaly detection algorithm, so as to make our clustering results more easily match the anomaly detection algorithm. In addition, evolutionary algorithms is used to configure appropriate detection processes and parameters for each cluster in large batches. After that, a complete online algorithm configuration process based on KPIs clustering is obtained, which can automatically match the appropriate anomaly detection algorithm for the new time series according to the results of the first two steps of offline training. The experiment proves that our cluster methods achieves the state-of-art results. The accuracy of our anomaly detection framework is 85.1%.